\documentclass[dvips]{imsart}
\usepackage[OT1]{fontenc}
\usepackage{amsthm,amsmath}
\usepackage{amsfonts}
\usepackage{amsbsy}
\usepackage{subfigure}
\usepackage{wrapfig}
\usepackage{amssymb}
\usepackage{latexsym}
\usepackage{euscript}
\usepackage{fancyhdr}
\usepackage{epsfig}
\usepackage{graphicx}
\usepackage{subfigure}
\usepackage{fancybox}
\usepackage{varioref}
\usepackage{enumerate}
\usepackage{verbatim}
\usepackage{makeidx}
\usepackage[english]{babel}
\usepackage[applemac]{inputenc}

\newcommand{\R}{\mathbb{R}}
\newcommand{\E}{\mathbb{E}}

\newcommand{\1}{{\sf \hspace*{0.9ex}}\rule{0.15ex}{1.6ex}\hspace*{-1ex} 1}

\newcommand{\Argmax}{Argmax}

\newcommand{\X}{\mathcal{X}}

\arxiv{math.ST/1002.3744}

\newtheorem{Lemme}{Lemma}
\newtheorem{proposition}{Proposition}
 \newtheorem{remarque}{Remark}
 \newtheorem{theoreme}{Theorem}
 \newtheorem{definition}{Definition}
 
 \newtheorem{corollaire}{Corollary}
 
  \newtheorem{assumption}{Assumption}
\begin{document}

\begin{frontmatter}
\title{Plugin procedure in segmentation and application to hyperspectral image segmentation}
\runtitle{Plugin procedure in segmentation}
\begin{aug}
\author{\fnms{Girard} \snm{Robin}\thanksref{t1}\ead[label=e1]{robin.girard@mines-paristech.fr}}
\address{Sophia Antipolis, France\\}
\runauthor{R. Girard}
\affiliation{Mines Paristech}
\end{aug}
\begin{abstract}
  In this article we give our contribution to the problem of segmentation with plug-in procedures. We give general sufficient conditions under which plug in procedure are efficient. We also give an algorithm that satisfy these conditions. We give an application of the used algorithm to hyperspectral images segmentation. Hyperspectral images are images that have both spatial and spectral coherence with thousands of spectral bands on each pixel. In the proposed procedure we combine a reduction dimension technique and a spatial regularisation technique. This regularisation is based on the mixlet modelisation of Kolaczyck and Al. \cite{mulgran}. 
\end{abstract}



\end{frontmatter}

\section{Introduction}
In this article we study the segmentation problem which is a particular learning problem that generalizes classification (as defined in \cite{esaimsurvey}) by asking for multiple simultaneous decisions instead of a simple decision. In segmentation, we have an observation $x=(x[1],\dots,x[N])$ in  $\X^N$ (in this paper, $\X=\R^p$ but some results are more general). This observation is associated to a label $y=(y[1],\dots,y[N])$ with values in the product space $\{0,1\}^N$. Note that we restrict ourself to the binary segmentation mainly to simplify the theoretical study, however, in the applications of this paper $y$ takes values in $\{1,\dots,K\}^N$. In the segmentation problem, the label $y$ is unknown and a segmentation procedure is a function $g:\X^N\rightarrow \{0,1\}^N$ that tries to guess the correct label associated to a given observation. For example, in a grayscale image segmentation, $N$ is the number of pixel in the image and $\X=\R$, in the hyper-spectral image segmentation problem $\X=\R^p$ with $p$ very large.  The segmentation error of a segmentation procedure $g$ can be measured with a distance $d$ on $\{0,1\}^N$ by $d(g(X),Y)$. In this article we will use the normalized Hamming distance $d_{H}$ defined by 
\[\forall x,y\in \{0,1\}^N \;\; d_{H}(x,y)=\frac{1}{N}\sum_{i=1}^N\1_{x\neq y}.\]
The value of $d_H(g(X),Y)$ represents the proportion of misclassified pixels. 
\\

In order to analyze the theoretical performances of the proposed procedures, we introduce a probabilistic setting, and let $(X,Y)$ be an $\X^N\times \{0,1\}^N$ valued random pair, modeling an observation and the corresponding label. Let $P_X$ be the distribution of $X$, $P_{Xi}$ the distribution of $X[i]$ for all $i\in \{1,\dots,N\}$, and $P_Y$ the distribution of $Y$. For $k=0,1$, $i=1,\dots,N$, let $P_{ik}$ be the probability distribution of $X[i]$ given $Y[i]=k$, let $\pi_i=P_Y(Y[i]=0)$.  The distribution of the random pair $(X,Y)$ may be described by $((P_{ik})_{i,k},(\pi_i)_i)$. In this article, we make the following assumption\\
\begin{assumption}\label{AA}
For all $i,j\in\{0,\dots,N\}$, $i\neq j$, the random variables $X[j]$ and $Y[i]$ are independent.\\
\end{assumption}

We measure the performance of a segmentation procedure $g$ by $\E[d_{H}(g(X),Y)]$ and it  is easy to see that, under assumption \ref{AA}, the optimal procedure, e.g the one that minimizes $\E[d_{H}(g(X),Y)]$, is given by 
\begin{equation}\label{optimal}
\forall i\in \{1,\dots, N\}\;\; g^*[i](X)=\1_{1<R_{\pi}[i]}\;\; R_{\pi}[i]=\frac{1-\pi_i}{\pi_i}\frac{dP_{i1}}{dP_{i0}}(X[i])
\end{equation}

\paragraph{The two step framework} in a plug-in perspective, the construction of a segmentation procedure approaching $g^*$ can be divided into two steps.
\begin{itemize}
\item \textbf{Step 1: Learning step}. Find the substitute $(\tilde{P}_{i0},\tilde{P}_{i1})_{i=1,\dots,N}$ for the conditional distributions on each pixel $(P_{i0},P_{i1})_{i=1,\dots,N}$. 
\item \textbf{Step 2: Segmentation Step}. Find $\hat{P}\in \times_{i=1}^N Conv(\tilde{P}_{i0},\tilde{P}_{i1})$ \footnote{$Conv(\tilde{P}_{i0},\tilde{P}_{i1})$ is the convex hull of \{ $\tilde{P}_{i0},\tilde{P}_{i1}\}$} using the observation $X$ drawn from $P_X$ (the observed image). Note that finding such a distribution is equivalent with finding a set of weights $\pi(\hat{P}_X)=(\pi_{i}(\hat{P}_X))_{i=1,\dots,N}\in [0,1]^N$ with 
\begin{equation}\label{hatpx}
\hat{P}_X=\prod_{i=1}^N\left ( \pi_{i} (\hat{P}_X)\tilde{P}_{i0}+(1-\pi_{i}(\hat{P}_X))\tilde{P}_{i1}\right ).
\end{equation}
The (plugin) segmentation procedure obtained with this construction is 

\begin{equation}\label{plugin}
\forall i\in \{1,\dots, N\}\;\; \hat{g}(X)[i]=\1_{1<\hat{R}_{\pi}[i]}\;\;\hat{R}_{\pi}=\frac{1-\pi_{i} (\hat{P}_X)}{\pi_{i} (\hat{P}_X)}\frac{d\tilde{P}_{i1}}{d\tilde{P}_{i0}}(X[i]).
\end{equation}
\end{itemize}

\begin{remarque}
The segmentation rule depends on the observation $X[i]$ through the evaluation of the likelihood ratio $\frac{d\tilde{P}_{i0}}{d\tilde{P}_{i1}}(X[i])$  but also depends on the whole image $X$ through the evaluation of the weigh vector $\pi(\hat{P}_X)$ in step $2$. 
\end{remarque}

In the applications of this article, a learning set, composed of $n$ independent random variables drawn from $P_{0i}$ and $P_{1i}$ $\forall i=1,\dots,N$, is given in the first step. This is what we will refer to as the supervised segmentation, and this justifies the name of the first step.

\paragraph{Obtained rate of convergence.} 
In order to measure the difference, between a segmentation procedure $g$ and the best segmentation procedure $g^*$, it convenient to introduce the excess risk:\begin{equation}\label{excess}
\mathcal{S}(g)=\E[d_{H}(g(X),Y)]-\E[d_{H}(g^*(X),Y)].
\end{equation}
 In this article, we give rates for the convergence of $\mathcal{S}(g)$ to zero. The procedure we describe in Section \ref{main} for the estimation of the weight $\pi(P_X)$ is a general model selection procedure introduced by Kolaczyk et Al. \cite{mulgran}. The algorithm in step $2$ has never been used before but it is in line with step (c) in the work of Antoniadis et Al. \cite{Antoniadis:2009rt}. Apart from our numerical studies and the fact that we combine step $1$ and step $2$, our main contribution is theoretical. Indeed, we obtained rates of convergence for a class of plugin segmentation procedures. The corresponding results are summarized in Theorem \ref{main}. It gives a relation between the convergence of $\mathcal{S}(g)$ and separately with the choice of $(\tilde{P}_{i0},\tilde{P}_{i1})_{i=1,\dots,N}$ and the complexity of the class of possible weights $\pi(P_X)$.\\  
  As an example, when $(P_{ik})_{i=1,\dots,N,k=0,1}$ are gaussian, the procedure
 $\hat{g}$ we give satisfies
\[\mathcal{S}(\hat{g}) \lesssim \sqrt{\frac{\log(N)}{N}}+\frac{log(p)}{\sqrt{n}},\]  

if 
\begin{itemize}
\item $(P_{ik})_{i=1,\dots,N,k=0,1}$ have the same covariance
\item $(P_{ik})_{i=1,\dots,N,k=0,1}$ have means satisfying a sparsity assumption. 
\item $f:i\rightarrow \pi(P_X)[i]$ satisfies some smoothness assumption (which are fulfilled in the case of the boundary fragment model as described in \cite{imagerecons}). 
\end{itemize}  
(recall that $p$ is the dimension of $\X$, $N$ is the number of pixels in the image, $n$ is the size of the learning set and the notation $A\lesssim B$ means that there exists a constant $c>0$ such that $A\leq c B$.)

\paragraph{Foreseen applications.} In satellite imagery, images often contain more than 200 spectral bands. In mars satellite imagery, (see \cite{Schmidt:2007sf}) geologists have a clear idea of what type component they will find within images and they can create a learning set. This learning set can be made out of samples from images that have been analyzed by an expert. Anyway, this expert cannot identify spectra from the tera bytes of data that flows from Mars to the earth, and the proportions of the different component in the learning set taken from a randomly chosen place on mars cannot be used to infer on what will be the proportion in a new image coming from another part of Mars.\\

In medical imagery of the brain the problem is also exploratory, but the number of images is relatively small and if experts can analyze images by themselves, contamination by noise makes a statistical support attractive. Images contain thousands of spectral bands. \\

 The remote-sensing literature about supervised and unsupervised segmentation procedure of images is really large, however, only a few procedures have been developed to tackle the multi and hyperspectral image segmentation problem.  Finaly, we are not aware of any work providing theoretical assessment of image (or hyperspectral image) segmentation procedure with a learning step.

\paragraph{Structure of the paper.} This article is constructed as follows. In Section \ref{main} we give our main theoretical result which concerns step $2$ (segmentation step). In Section \ref{dimred} we give an algorithm that aims at estimating the conditional density under gaussian assumption and when $\X$ is a high dimensional space $\R^p$ with $p$ large. This algorithm gives a solution for step $1$ and is shown to satisfy necessary condition for step $2$ to be consistent (i.e Assumptions \ref{chideux}  from Section \ref{main}). In Section \ref{applic} we apply the whole algorithm (step $1$ plus step $2$) to hyperspectral (medical and satellite) image segmentation. In Section \ref{sec:Thres} we give the proof of our theoretical results. 

\section{Algorithm and main result}\label{main}
In this section, we give the algorithm for step $2$ (estimating the weights) and associated theoretical results. This can be considered as the main result of this paper. 

\subsection{Mixlet estimation}

The mixlet algorithm has been introduced by Kolaczyk et Al.  \cite{mulgran}. It a model selection algorithm based on a penalized maximum likelihood estimation.\\

Let $\mathcal{M}_N$ be a finite subset of $\times_{i=1}^N Conv(\tilde{P}_{i0},\tilde{P}_{i1})$ (i.e a subset of models) and $pen_N : \mathcal{M}_N\rightarrow \R_+$ a penalty function. Note that the set $\mathcal{M}_N$ can either be seen as a set of measures, as a set of weights, or (because it is finite) as a set of densities with respect to a dominating probability measure. The mixlet estimation of $\pi(P_X)$ is obtained by finding $\hat{P}_X$ given by 
\begin{equation}\label{modelalgo}
\hat{P}_X=\Argmax_{Q\in \mathcal{M}_N}\left \{log(Q(X))-4pen_N(Q) \right \}.
\end{equation}

For the penalty function and the associated set of models, we only require a kraft inequality:

 \begin{assumption}\label{modelselc}
 The set of models $\mathcal{M}_N$ and the associated penalty function $pen_N$ satisfy 
 \[\sum_{Q\in \mathcal{M}_N}e^{-pen_N(Q)}\leq  1\]
 for a positive constant $C$.
 \end{assumption}

This type of assumption is standard in model selection theory (see \cite{Barron:1999og}). This can be seen as a complexity assumption on the set of models and penalty.   Such inequality then results from Kraft inequality, as, for example in Kolaczyk and Nowak \cite{multlik}. Equivalently, it can be seen as a topological covering bound, as, for example in Barron et Al. \cite{Barron:1999og}. The way we use this inequality in the proof of the following theorem is inspired from the work of Birgé \cite{Birge:2003rn}.

\subsection{The "d-dimensional" hyperspectral image}
In this example, we examine the choice of $\mathcal{M}_N$ and $pen_N$ in a particular case related to a "d-dimensional" hyperspectral image.   \\

 Each index of $\{1,\dots,N\}$ will now be associated to the center of one of the $N=n^d$ pixels of a $d$-dimensional hypercube: $[0,1]^d$ (here we assume that pixels in the image are d-dimensional hypercube with equal size). As a consequence, giving a segmentation $g(X)$ of $X$ is equivalent with finding a particular partition of $[0,1]^d$ into groups of pixels. We will search those partition within the set of recursive dyadic partition. Recall that a recursive dyadic partition of $[0,1]^d$ (in short $RDP_d$) is a partition constructed recursively and associated to a $2^{d}$-tree, e.g a tree with $2^{d}$ sons or one leave at each node. A splitting of a node in the tree correspond to a splitting of a $d$-dimensional hypercube into $2^d$ identical hypercubes. \\

The set of models $\mathcal{M}_N$ and the associated penalty function will be use in the numerical application to hyperspectral image segmentation. For $i=1,\dots,N$, we will search $\pi(\hat{P}_X)_i$ in a regular grid of $[0,1]$ with $N^{3/2}$ elements (take the entire part of $N^{3/2}$ if it is not round). This grid of $[0,1]$ will be denoted $G_N$. Finally $\mathcal{M}_N$ will be the set of product distribution $Q=\Pi_{i=1}^NQ_{i}$ on $\X^N$ with $Q_i\in conv(\{\tilde{P}_{i1},\tilde{P}_{i0}\})$, $\pi(Q)_i\in G_N$ for all $i=1,\dots, N$ and $i\rightarrow \pi(Q)_i$ constant on each piece of a given $RDP_d$. The minimal $RDP_d$ on which $\pi(Q)_i$ is constant will be $\mathcal{P}(Q)$.
The function $pen(Q)$ penalize the partitions that are too rich:
\begin{equation}
pen(Q)=m^{d-1}\left (\frac{3}{2}\log N+\frac{4}{3}\log 2\right ),
\end{equation}
where $m^{d-1}$ is the number of elements of the partition $\mathcal{P}(Q)$. It is known that with this type of penalty, we have a kraft inequality
 \begin{equation}\label{equ:kraft}
 \sum_{Q\in \mathcal{M}_N}e^{-pen(Q)}\leq 1,
 \end{equation}
 (see for example \cite{mulgran}) end hence Assumption \ref{modelselc} is fulfilled.\\
 
The corresponding model selection algorithm (for step $2$), i.e used to find the minimum in Equation (\ref{modelalgo}) with the defined set of models and associated penalty function, can be implemented efficiently with a pyramidal algorithm (see \cite{mulgran}) and has been called mixlet algorithm.

\subsection{Theoretical result}
Before we state the theoretical result we give assumptions that have to be fulfilled
\begin{assumption}\label{deux}
 There exists a positive constant $c$ such that $\inf_{1\leq i \leq N}|P_{i0}-P_{i1}|_1\geq c$, where $|P-Q|_1$ is the $L_1$ distance between $P$ and $Q$ (two distributions on $\X$) given by $|P-Q|_1=\int_{\X}|dP-dQ|$.
\end{assumption}
\begin{assumption}\label{BB}
 There exists a positive constant $c'$ such that $\inf_{1\leq i \leq N}\min(\pi_{i},1-\pi_i)\geq c$.
\end{assumption}
\begin{assumption}\label{chideux}
There exists $C>0$ such that 
\[\forall k_1,k_2\in\{0,1\}, \;i\in \{1,\dots,N\}\;\; \chi^{2}(P_{ik_1},P_{ik_2})<C, \]
where $\chi^2(P_1,P_0)$ is the chi square divergence between two probability distribution ($P_1$ and $P_0$)  defined by 
\begin{equation}\label{chisquare}
\chi^2(P_1,P_0)=\left \{\begin{array}{cc}\int \left (\frac{dP_1}{dP_0}-1\right )^2dP_0 & \text{ if } P_1 \ll P_0\\ \infty & \text{ else } \end{array}\right. 
\end{equation}
\end{assumption}

The obtained result is the following

\begin{theoreme}\label{Mainth}
Under the Assumptions [\ref{AA}-\ref{chideux}],  and if $\hat{g}$  is a classification rule  constructed with the two step given in the preceding section (defined by Equation (\ref{plugin})), we have
\begin{equation}
N\mathcal{S}(\hat{g})\lesssim \psi_{N,n}=\mathcal{L}_{N,n}+\inf_{R\in \mathcal{M}_N}h(R,N) 
\end{equation}
as long as $N e^{-N\psi_{N,n} }=O(\psi_{N,n} )$ where $c_0$ is a positive constant, $\mathcal{S}$ the excess risk defined by Equation (\ref{excess}) 
\begin{equation}\label{gRN}
\forall R\in \mathcal{M}_N,\;\;  h(R,N)=\|\pi(P_X)-\pi(R)\|_{l_1}+pen_N(R).`
\end{equation}
and the error term related to the learning step is given by 
\begin{equation}\label{equ:term2}
\mathcal{L}_{N,n}=\sum_{i=1}^N\E\left [\Omega^2\left (\frac{dP_{0i}}{dP_{1i}}\frac{d\tilde{P}_{1i}}{d\tilde{P}_{0i}}\right )\right ]+D_{i0}+D_{i1}
\end{equation}
where 
\[D_{i0}=\max\left ( \chi^2(P_{i0},\tilde{P}_{i0}),\E_{P_{i0}}\left [\frac{P_{i1}-\tilde{P}_{i1}}{\tilde{P}_{i1}}\right ]\right )\]
and 
\[D_{i1}=\max\left ( \chi^2(P_{i1},\tilde{P}_{i1}), \E_{P_{i1}}\left [\frac{P_{i0}-\tilde{P}_{i0}}{\tilde{P}_{i0}}\right ] \right ).\]
\end{theoreme}

The proof of this result is postponed in the annex. Let us discuss the assumptions of this Theorem. 
\begin{itemize}
\item In order to get a rate of convergence for the full process (step $1$ and step $2$) we need to provide a bound to $\E[\mathcal{L}_{n,N}]$ (where this last expectation is with respect to the learning set). This will be the purpose of Section \ref{dimred}.

\item Assumption \ref{BB} is rather strong. If the number of pixel with a pure component ($\pi_i=0$ or $1$) is small (i.e the order of the upper bound in the preceding theorem), the results are still valid. We think that the construction of $\mathcal{M}_N$ should be changed to avoid this assumption, in particular the discretization of the set of values for $\pi_i$ should be refined near $0$ and $1$. This will be the purpose of further work.  
\item Assumption 4 is related to the choice of the model in ad-equation  with the structure of $\pi(\hat{P}_X)$. In the next section we explain how this choice can be done in the case of a "$d-$dimensional image". 
\end{itemize}

\subsection{Turning back to the "d dimensional image"}\label{sec:turning}

In order to be able to upper bound $\inf_{R\in \mathcal{M}_N}h(R,N)$ (tradeoff between bias and complexity) it is natural to introduce assumption about the "spatial" regularity (i.e regularity of the weights in the image) that can be handled by a $RDP_d$. This is done by the following Definition and Assumption

\begin{definition}\label{def:CM}
Let $f:[0,1]^d\rightarrow \R$ be a piecewise constant function and $B(f)$ be the set of points on which $f$ is not continuous. Let $N(f,r)$ be the minimal number of hypercubes from an RDP with lenght $r$ that cover $B(f)$. To each $\beta>0,M>0$ we associate the set $CM_d(\beta,M)$ of piecewise constant functions defined by
  \[\left \{f:[0,1]^d\rightarrow \R\;:\; f \text{ piecewise constant, }\|f\|_{\infty}\leq M\;\; \text{ and }N(f,r)\leq \beta r^{-(d-1)}\right \}.\]
  \end{definition}
 \begin{assumption}\label{regularity}( $d$-dimensional regular image, $d\geq 2$).  Let $PI_N$ be the regular partition of $[0,1]^d$ into $N$ identical hypercubes (i.e $N$ pixels). For all $k\in \{1,\dots,K\}$, there exists $\beta>0$, $M<\infty$ and $f_k\in CM_d(\beta,M)$ (see definition (\ref{def:CM})) such that 
\begin{equation}
\forall i\in \mathcal{T}_N,\;\;\; \pi_{ik}=f_k(t_i), 
\end{equation}
where $t_i$ is the center of the hypercube $i$ of $PI_d$. 
\end{assumption}
\begin{remarque}
This assumption is an assumption on the topological structure of $\mathcal{T}_N$. This structure is more complex when $d$ is bigger.
\end{remarque}
\begin{proposition}\label{proposition1}
With $\mathcal{M}_N$ and $pen_N$ as defined previously, and under Assumption \ref{regularity}, we have
\[\inf_{R\in \mathcal{M}_N}h(R,N)\leq c N \left (\frac{\log(N)}{N}\right )^{1/d}\]
for a positive constant $c$.
\end{proposition}
The proof of this proposition can be founded in Donoho \cite{wedg} or in the Annex of Kolaczyk et Al.  \cite{mulgran}.

\begin{corollaire}\label{segmenTH}
Let $d\geq 2$ and suppose that for all $i=1,\dots,N$, $k=0,1$, we have $\tilde{P}_{ik}=P_{ik}$. Under the Assumptions \ref{AA}, \ref{deux}, \ref{BB} and if $\hat{g}$  is a classification rule  constructed with the two step given in the preceding section (defined by Equation (\ref{plugin})) with $\hat{P}_{X}$ in the first step, given by Equation (\ref{modelalgo}), and ($\mathcal{M}_N$,$pen_N$) as defined in this section with  Assumption \ref{regularity} fulfilled, then there exists a positive $c_0$ such that
\[
\mathcal{S}(\hat{g})\leq c_0\left (\frac{\log(N)}{N}\right )^{1/d},
\]
where $\mathcal{S}$ is the excess risk of segmentation refined by (\ref{excess}).
\end{corollaire}
This corollary is a direct consequence of the preceding theorems. 
\begin{remarque}
This result together with the one in the next Section may be seen as a complete convergence description of the algorithm that is used in Section \ref{applic}. Unfortunately it is not the case because the only application we have are not in the case where the number of possible class is two.\\
\end{remarque}


\section{Handling the learning step}\label{dimred} 
\subsection{Dimension reduction in segmentation: a solution to step $1$ in high dimension}

In this section, we investigate Step $1$ when $\X=\R^p$ under the following assumption
\begin{assumption}\label{a:gaussian}
For $k=0,1$, $i=1,\dots,N$, $P_{ki}$ is gaussian with mean $\mu_{k}$ and covariance $C$. For $k=0,1$,  $C^{-}\mu_k$ has less than $p_0+1$ non null components, where $p_0$ is bounded with respect to $p$, $n$ and $N$. The matrix $C$ is diagonal. 
\end{assumption}
Note that, under this assumption, $P_{ki}$ does not depend on the position $i$. 

 For $k=0,1$, we suppose that we have $n_k$ independent random variables $Z_{kj}$ ($n_k$ is a positive integer) drawn from distribution $P_{k1}$.  The set $\mathcal{Z}=\{Z_{kj},\;\; k=0,1,\;j=1,\dots,n_k\}$ is the learning set.  If $A$ is a squared matrix, we will use the notation $A^-$ for the associated generalised inverse.\\
 
 The algorithm for estimating $P_{k1}$ ($k=0,1$), i.e the learning step, is as follows.

\begin{enumerate}
\item For $i=1,\dots,p$, compute $\bar{\sigma}^2[i]$ the unbiased empirical variance of $(Z_{kj}[i])_{k=0,1,j=1,\dots,n_k}$ and for $k=0,1$ compute $\bar{\mu}_k$ the empirical mean of $(Z_{kj})_{j=1,\dots,n_k}$.\\
\item Compute  $\hat{I}$ as 
\[\hat{I}=\bigcup_{k=0,1} \left \{i\in\{1,\dots,p\}; :\; \frac{|\bar{\mu}_k[i]|}{\bar{\sigma}[i]}>\sqrt{2\frac{\log(p)}{n}} \right \} \]
\item The means $\mu_0$ and $\mu_1$ are estimated by 
\[\hat{\mu}_k[i]=\left\{\begin{array}{cc} \bar{\mu}_{k}[i] & \text{ if } i\in \hat{I}  \\0 & \text{ else }\end{array}\right. i=1,\dots,p,\; k=0,1,\]
and the covariance $C$ by the diagonal matrix $\hat{C}$ with diagonal elements 
\[\hat{\sigma}[i]=\left\{\begin{array}{cc} \bar{\sigma}_{k}[i] & \text{ if } i\in \hat{I}  \\0 & \text{ else }\end{array}\right. i=1,\dots,p\]
\end{enumerate}
 
\begin{theoreme}\label{Th:last}
Let us take $\mathcal{M}_N,pen_N$ as in subsection \ref{sec:turning} with Assumption \ref{regularity} fullfiled. Let us make Assumption \ref{a:gaussian} and for $i=1,\dots,N$ $k=0,1$ compute $\tilde{P}_{ki}$ as a gaussian distribution with mean $\hat{\mu}_k$ but assuming the covariance matrix $C$ is known. Then under assumption \ref{AA}, \ref{deux}, \ref{BB} and if $\hat{g}$  is a classification rule  defined by Equation (\ref{plugin}) with $\hat{P}_{X}$ in the first step, given by Equation (\ref{modelalgo}),
we have:
\[\E[S(\hat{g})]\lesssim \sqrt{\frac{\log(N)}{N}}+\frac{log(p)}{\sqrt{n}}.\] 
\end{theoreme}
The proof of this Theorem is given in Subsection \ref{Proof:LastTh}.
The weakness of this theorem is that it require the knowledge of $C$. We did not succeed in giving a proof in more general case and we believe that further improvement of this result is beyond the scope of this paper.

\section{Application to hyperspectral image segmentation}\label{applic}

Before we give the details of our application to hyperspectral medical image segmentation we have to emphasis that the theoretical results we gave are designed for a two class segmentation ($K=2$). However, in most application the number of possible classes is larger than two and the algorithms we gave can easily be extended to the case when $K>2$. Indeed, the penalized maximum likelihood estimation of the weight $(\pi_{i})_{i}$ can be used when $K>2$ and the dimension reduction algorithm can be extended to a multiclass framework. This last extension can be done using a global measure of the contrast between groups. 

\subsection{Application to medical hyperspectral segmentation}

\begin{figure}[htbp]
\begin{center}
\includegraphics[width=6cm,height=6cm]{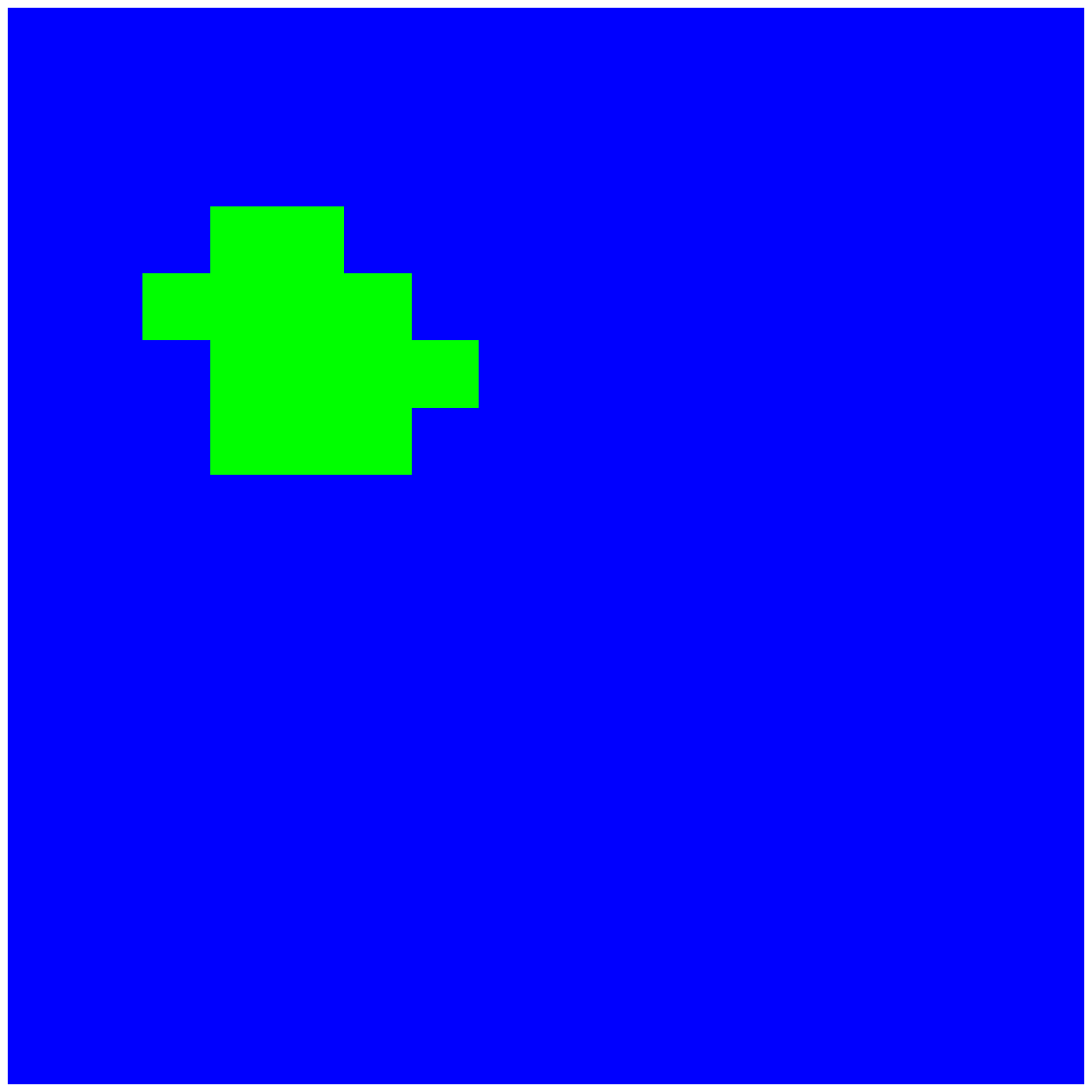}
\includegraphics[width=6cm,height=6cm]{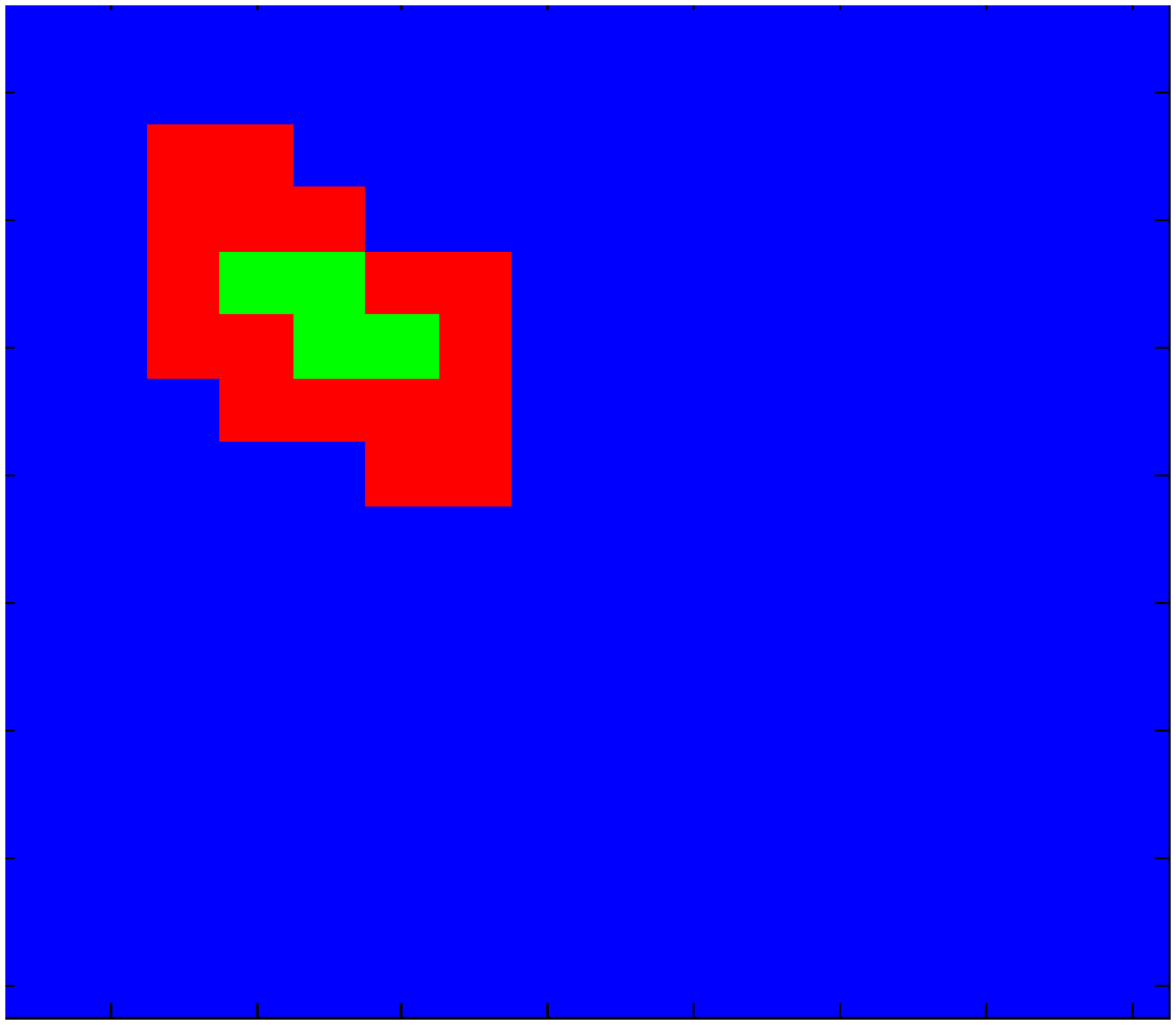}
\caption{Obtained segmentation -on the left-, and segmentation that we should obtain -on the right-. The pixels that are colored in blue correspond to healthy tissues, the green is associated to type B Glioblastomas and the red is associated to type A glioblastomas.}
\label{fig:tumeur}
\end{center}
\end{figure}

Hyperspectral images of the brain from magnetic resonance imaging are high dimensional data. These images have only a few pixel ($N=256$ pixels) giving the detail of a slice of the brain but on each pixel, we observe a high dimensional spectra with $p=1024$, hence $\X=\R^{1024}$. A given spectra is expected to give a complete information on the tissular characteristic at a given spatial position. These tissular characteristics can be classified into groups.  
In this medical problem, we have a learning set composed of $62$ spectra from four different groups: $21$ Glioblastomas of type $A$, $9$ Glioblastomes of type $B$, $16$ Méningiomes, and $9$ healthy tissues. We were given an hyperspectral image associated to a Glioblastoma mixing type A and type B. This image (its spatial configuration) is simulated from spectra obtained in a real experimentation. The obtained segmentation and the true segmentation are given in Figure \ref{fig:tumeur}. Our conclusion is that the tumor is well localized but that the different types of Glioblastomas are not distinguished. \\
\begin{figure}[htbp]
\begin{center}
\includegraphics[width=11cm,height=11cm]{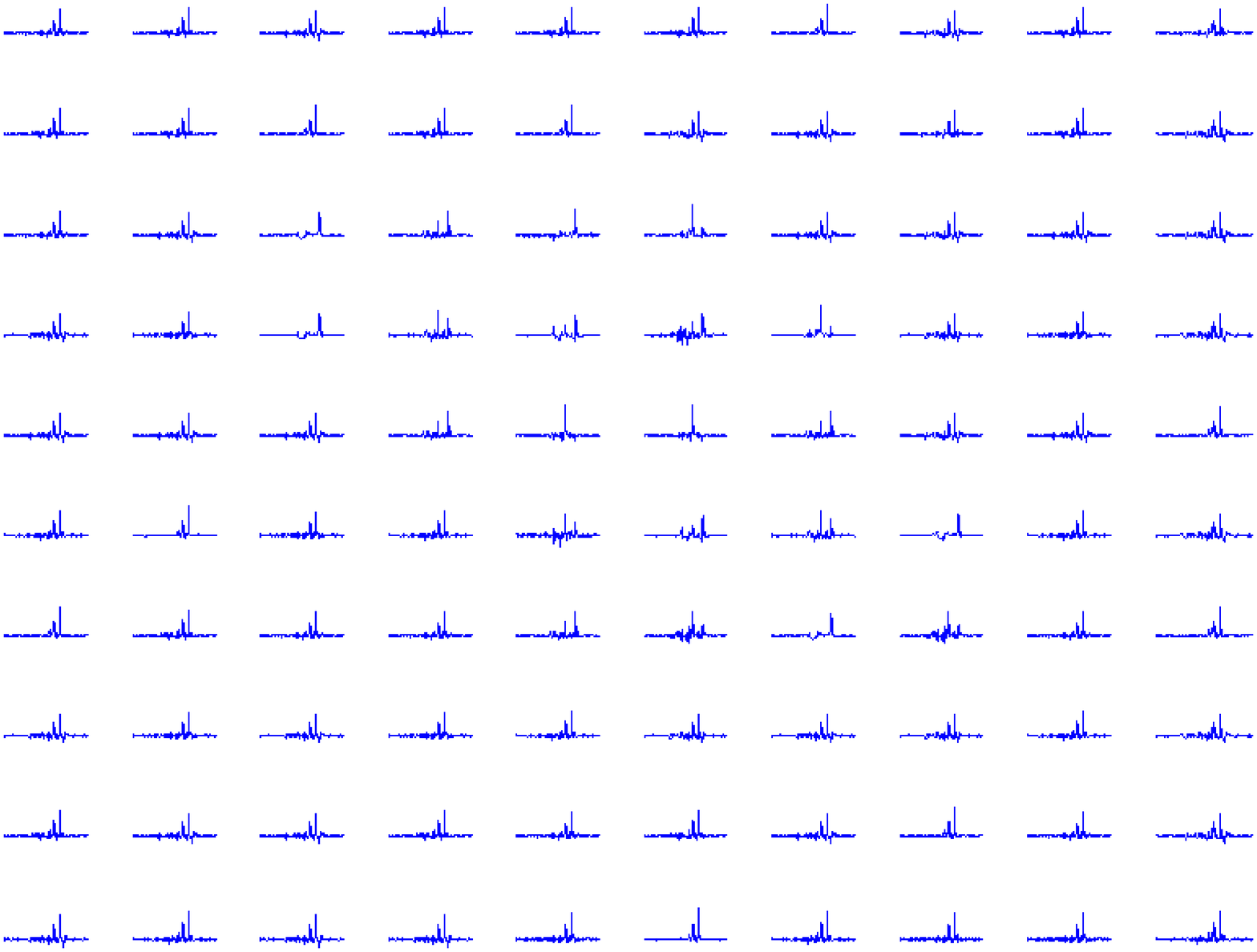}
\caption{ $10\times10$ square in the top left of the hyperspectral $16\times 16$ image of a glioblastoma.}
\label{fig:tumeur0}
\end{center}
\end{figure}

Note that if the result are positive, this partly results from a pre-treatment of the data (ad-hoc re-phasing of the spectra) and from the fact that we did not include any metastases in the problem (metastases and glioblastomas are hard to distinguish). Studying automatic re-phasing will be the purpose of further research.   

\section{Conclusion}
We studied the problem of supervised segmentation. We gave theoretical results in a plugin perspective that allow to consider a wide range of model selection procedure. We showed that the procedure of Kolaczyk et Al. \cite{mulgran} (the mixlet procedure) can be applied consistently for segmentation of images with smooth boundaries. We gave a theoretical result that separate the segmentation error due to the learning step and the segmentation error due to the segmentation step (estimation of the weigh in the mixture model). We gave a reduction dimension procedure for the learning step and gave associated theoretical results. The corresponding result gives the convergence rate of the whole segmentation procedure (learning step plus segmentation step), this convergence rate is adapted to the case when the dimension $p$ of the feature space of a pixel observation is much larger than the number $n$ of elements in the learning set. Finally, we applied the whole methodology to medical image segmentation. 

\section{Theoretical results}\label{sec:Thres}
\subsection{A general result in segmentation with a plugin rule}
  Any segmentation procedure $\hat{g}$ can be summarized by $\hat{R}_{\pi}:\X^N\rightarrow [0;\infty]^N$ through $\hat{g}[i](X)=\1_{1<\hat{R}_{\pi}[i](X)}$ $\forall i\in \{1,\dots,N\}$. We obtained Theorem \ref{genth} below which gives an upper bound on the excess risk $\mathcal{S}(\hat{g})$ under the following assumption on the error made while estimating $R_{\pi}$:

\begin{assumption}\label{assump:eta}
There exists $c_0,c_1,c_2>0$, a sequence $\psi_N$, with $N e^{-c' N\psi_{N} }=O(\psi_{N} )$ for any $c'>0$, such that $\hat{R}_{\pi}:\X^N\rightarrow [0,\infty]^N$, and $R_{\pi}=(R_{\pi}[i])_{i=1,\dots,N}$ satisfy 
 \begin{equation}\label{inq:expBirge}
 \forall \delta\geq 0  \;\;  P_X\left (\mathcal{E}(R_{\pi},\hat{R}_{\pi}) \geq \delta \right )\leq c_2e^{c_0N\psi_{N}-c_1\delta},
 \end{equation}
where $\mathcal{E}(R_{\pi},\hat{R}_{\pi})$ is given by
\begin{equation}\label{Er:eta}
\forall x,y\geq 0 \;\; \mathcal{E}(x,y)=\sum_{i=1}^N\Omega^2(x/y).
\end{equation}
and 
 \begin{equation}\label{Omega}
 \forall x\geq 0\;\; \Omega(x)=\frac{|x-1|}{x+1}.
 \end{equation}
 \end{assumption}
We also need the following additional assumption 
\begin{assumption}\label{assump:low}
There exists $c>0$ such that 

\begin{equation}\label{Er:low}
\forall i\in \{1,\dots,N\}\;\; \E[\Omega(R_{\pi}[i])]\geq c 
\end{equation}

 \end{assumption}
 
\begin{theoreme}\label{genth}
Let us take $\hat{R}_{\pi}:\X^N\rightarrow [0,\infty]^N$,  and $\hat{g}$ the associated segmentation procedure with, for all $i\in \{1,\dots,N\}$ $\hat{g}[i]=\1_{\hat{R}_{\pi}[i]>1}$. Take $R_{\pi}$ and the associated optimal segmentation procedure $g^*$ as defined by 
Equation \ref{optimal}.  Under Assumption \ref{AA}, and if $\hat{R}_{\pi}$ satisfies assumption \ref{assump:eta} \ref{assump:low}, then 
\[\mathcal{S}(g) \lesssim \psi_{N}\]
where $\mathcal{S}$ is the excess risk defined by Equation \ref{excess}.
\end{theoreme}

This Theorem is proven in Annex. Assumption \ref{assump:low} is a weak assumption that will be satisfied if $P_{0i}$ and $P_{1i}$ are not as closed as desired (in $i\in \{1,\dots, N\}$) for a well chosen distance. Notice that Assumptions \ref{deux} and \ref{BB} imply Assumption \ref{assump:low}.\\
The way to obtain inequality \ref{inq:expBirge} in Assumption \ref{assump:eta} will be the topic of Subsection \ref{sub:weights} but the Theory developed by Birgé in \cite{Birge:2003rn} is our main reference and inspiration on the topic.\\

To understand the interest of Theorem \ref{genth} one should notice that a simple analysis gives 
\begin{equation}\label{notgood}
\mathcal{S}(g)\leq \sum_{i=1}^N\E[\Omega(R_{\pi}/\hat{R}_{\pi})].
\end{equation}
On the other hand, it is possible (using same argument that are used in the proof of Corollary \ref{cro:birge}) to show that Assumption \ref{assump:eta} implies 
\[\E[\Omega^2(R_{\pi}/\hat{R}_{\pi})]\lesssim \psi_N\]
which gives
\[\mathcal{S}(g) \lesssim \sqrt{\psi_{N}}.\]
However, Assumption \ref{assump:eta} is weaker than 
\[\E[\Omega(R_{\pi}/\hat{R}_{\pi})]\lesssim \psi_N\]
and this shows that Theorem \ref{genth} is sharper than Equation \ref{notgood}.
\begin{proof}  
To simplify notation we will set
 \begin{equation}\label{artyp}
 M=\frac{1}{N}\sum_{i=1}\Omega(R_{\pi}[i])U_i
 ,\;\;DC=\sum_{i=1}^NU_i,\;
 \end{equation}
 \[\;\forall I\subset\{1,\dots,N\}\; c(I)=\frac{1}{|I|}\sum_{i\in I} \E[\Omega(R_{\pi}[i])], \text{ and }M^k=M1_{DC=k}.\]

The proof of the Theorem is decomposed into $3$ steps \\

\textbf{Step 1}: we claim that there exists $1>c>0$ such that
\begin{equation}\label{equ:claim1}
 P\left ( \sup_{|I|=k}\sum_{i\in I}\Omega(R_{\pi}[i])U_i\leq c k\right )\leq e^{-2ck}.
 \end{equation}

 (where the supremum is taken over all subset of $\{0,\dots,N\}$ of size $k$). Indeed, we can notice that we have, for all $I\subset \{1,\dots,N\}$ of cardinal $k$
 
 \begin{align*}
P\left ( \sum_{i\in I} \Omega(R_{\pi}[i])\leq c(I)k/2 \right )&\leq P\left ( \sum_{i\in I} \Omega(R_{\pi}[i])-\sum_{i\in I} \E[\Omega(R_{\pi}[i])] \leq -c(I)k/2 \right )\\
&\leq e^{-\frac{c^2(I)k}{2}} 
 \end{align*} 
 where this last inequality results from the bounded difference inequality. Also, setting $\inf_{I: |I|=k}c(I)=2c$ ($c>0$ from Assumption \ref{troisbis}) gives Inequality \ref{equ:claim1}. \\

\textbf{Step 2}: we claim that with $c_0,c_1>0$ and $\psi_{N,M}$ as in Assumption \ref{assump:eta} we have

\begin{equation}\label{equ:claim2}
 P\left (\sup_{|I|=k}\sum_{i\in I}\Omega(R_{\pi}[i])U_i\geq c k\right )\leq e^{c_0N\psi_{N,n}-c_1c^2 k}.
 \end{equation}
Cauchy Schwartz inequality gives, for all $I\subset \{1,\dots,N\}$ of cardinal $k$, 
\begin{align*}
 \left ( \sum_{i\in I}\Omega(R_{\pi}[i])U_i\right )^2&\leq \left (\sum_{i\in I}\Omega(R_{\pi}[i])U_i\right )^2\\
  &\leq k\sum_{i=1}^N\Omega^2(R_{\pi}[i])U_i.
\end{align*}
which implies:
\[
 P\left (\sup_{|I|=k}\sum_{i\in I}\Omega(R_{\pi}[i])U_i\geq c k \right)\leq P_X\left (\sum_{i=1}^N\Omega^2(R_{\pi}[i])U_i\geq c^2 k\right ). 
\]
Finally, Inequality \ref{equ:claim2} follows from Assumption \ref{assump:eta} and the fact that, for any $i\in \{1,\dots,N\}$, $U_i=1$ implies that $\Omega(R_{\pi}[i])\leq \Omega(R_{\pi}[i]/\hat{R}_{\pi}[i])$.\\
 
\textbf{Step 3}: Standard calculous (see for example \cite{Devroye:1996fk} noticing that $\Omega(R_{\pi}[i])=|\eta_i-1/2|$) leads to 
 \[ \E\left [\sum_{i=1}^N1_{\hat{g}(X)[i]\neq Y_i}-1_{g^*(X)[i]\neq Y_i}|X\right ]=2M.\]
Also, we have
\begin{align*}
\E[M]& \leq\frac{c_0+1}{c_1c^2}+\sum_{N \geq k \geq \frac{c_0+1}{c_1c^2}N\psi_{N,n}}M^k\\
 & \leq \frac{c_0+1}{c_1c^2} N\psi_{N,n} +\sum_{N \geq k \geq \frac{c_0+1}{c_1c^2}N\psi_{N,n}} \sup_{|I|=k}\sum_{i\in I}\Omega(R_{\pi}[i]/\hat{R}_{\pi}[i]) \\
& \leq \frac{c_0+1}{c_1c^2} N\psi_{N,n} + \sum_{N \geq k \geq \frac{c_0+1}{c_1c^2}N\psi_{N,n}} k e^{c_0N\psi_{N,n}- k}+k  e^{-2c\frac{c_0+1}{c_1c^2}k}\\
&\leq \frac{c_0+1}{c_1c^2} N\psi_{N,n} + N^2 e^{- N\psi_{N,n} }+N^2  e^{-2\frac{c_0+1}{c_1c}N\psi_{N,n}}
\end{align*}
where these last two inequalities follows from the results of step $1$ and $2$. Since $N e^{-c' N\psi_{N,n} }=O(\psi_{N,n} )$ for any $c'>0$ this gives the desired result
 \end{proof}

\subsection{A general oracle inequality for penalized maximum likelihood estimation in mixlet model}\label{sub:weights}
In this Subsection, we give a general result on the estimation of the weigths. 
We first define the mean Hellinger distance between product measures. 
\begin{definition}
 If $P=\Pi_{i=1}^NP_i$ and $Q=\Pi_{i=1}^NQ_i$  are two product distributions on $\X^N$, we will call mean Hellinger distance: $H_N$, the positive quantity defined by  
 \begin{equation}
 H_N(P,Q)^2=\frac{1}{N}\sum_{i=1}^Nh^2(P_i,Q_i),
 \end{equation}
 where $h^2(P_i,Q_i)=\int_{\X}(\sqrt{dP_i}-\sqrt{dQ_i})^2$ is the squared Hellinger distance between $P_i$ and $Q_i$.
 \end{definition}
 

\begin{theoreme}\label{expth}
Suppose that $\mathcal{M}_N$ satisfies the Assumption \ref{modelselc} and that $\hat{P}_X$ is given by Equation \ref{modelalgo}. Then, under Assumption  \ref{chideux}, there exists $c',c''>0$ such that 
\[\forall \delta\geq 0  \;\; P\left (NH^2_N(P_X,\hat{P}_X)\geq \delta\right )\]
\[\;\;\leq \exp\left \{ c' \inf_{R\in \mathcal{M}_N}\{h(R,N)\}+c'' \mathcal{L}_{N,n} -\delta/4\right \},\]
where $h$  is given by Equation \ref{gRN} and $\mathcal{L}_{N,n}$ by Equation \ref{equ:term2}.
\end{theoreme}

Before we give the proof of this theorem, let us give some comments. The result of this theorem is an oracle inequality aiming to verify assumption \ref{quatre} as it has been noticed in Subsection \ref{sec:turning}. The function $g(R,N)$ measures the tradeoff between bias and complexity for model $R$. The error term $\sum_{i=1}^N\chi^2(\tilde{P}_{i1},P_{i1})+\chi^2(\tilde{P}_{i0},P_{i0})$ is related to step $1$ but should also be connected to Remark \ref{remdraw}. \\

 The Assumption \ref{chideux} is necessary to obtain theoretical results. It is weaker than the Assumption
  \begin{equation}\label{usual}
\sup_{x\in \X, k_1,k_2\in \{1,\dots,K\}^2}\frac{dP_{k_1}}{dP_{k_2}}(x)\leq B
  \end{equation}
  which is common in mixture model estimation (see for example the thesis of Li \cite{Li:1999fk}, or the work of Kolaczyk et Al.  \cite{mulgran}). Note that the Assumption given by (\ref{usual}) is not satisfied when the $P_{k}$ are gaussian.  Our Assumption \ref{chideux} allows us to consider gaussian mixture.

  Kolaczyk et Al. \cite{mulgran} introduced the idea of mixture weight estimation by maximum likelihood estimation. In the same paper, they give a theoretical result without using the mean Hellinger distance which weakened their result. In addition, they consider only the case where $\tilde{P}_{ik}=P_{ik}$ with $d=2$ and use the assumption related to Equation \ref{usual}. From this point of view, our result (Theorem \ref{expth} together with Proposition \ref{proposition1})  is a significant improvement of the result obtained in \cite{mulgran}. Indeed, for all $i=1,\dots,N$ $k=0,1$,  under assumption given by equation (\ref{usual}), and assumption \ref{regularity} with $d=2$, there exists a positive constant $c_0$ such that 
\[\E\left  [H_N^2(P_X,\hat{P})\right ]\leq c_0\left (\frac{\log N}{N}\right )^{1/2}.\]
We did not succeed in using this last bound to obtain a result in the segmentation problem (such as Theorem \ref{Mainth}). This is the reason why we worked on obtaining  stronger results such as Theorem \ref{expth} and its consequences.

The result may be difficult to apprehend in the preceding theorem, also we give the following simple corollary (it is a weaker result).

\begin{corollaire}\label{cro:birge}
Let $q\geq 1$. Under the assumption  of the preceding theorem, there exists a positive $c_0$ such that
\[\E[H^{2q}_N(\hat{P},P_X)]\leq c_0\frac{1}{N^{q}}\left ( \inf_{R\in \mathcal{M}_N}\{h(R,N)\}+  \mathcal{L}_{N,n}\right )^{q}.\]
\end{corollaire}

\paragraph{Proof of Theorem \ref{expth}.}
 
\begin{proof}
The proof relies on the same principle that the one exposed by Birgé in \cite{Birge:2003rn}. More precisely, the density $\hat{P}_X$ is a penalized maximum likelihood estimator, but it is also a $T$-estimator.  
As a consequence, we have
\begin{equation}\label{debu6}
P_{X}(NH_N^2(P_X,\hat{P}_X)\geq \delta) \hspace{7cm} 
\end{equation}
\[\hspace{0,5cm}\leq P_X\left (\exists Q\in \mathcal{M}_N\;: NH_N^2(P_X,Q)\geq \delta\right .\hspace{6cm} \]
\[\hspace{1cm} \left . \text{ and }\forall R\in \mathcal{M}_N \;\; \log \frac{Q(X)}{R(X)}\geq 4(pen_N(Q)-pen_N(R))\right ).\]
\[\hspace{0,5cm}\leq \sum_{Q\in \mathcal{M}_N}P_X\left (NH_N^2(P_X,Q)\geq \delta \right . \hspace{7cm} \]
\[\hspace{1cm} \left .\text{ and }\forall R\in \mathcal{M}_N \;\; \log \frac{Q(X)}{R(X)}\geq 4(pen_N(Q)-pen_N(R))\right )\]
(from the sub-additivity of probability measures).
In addition, for all $Q\in \mathcal{M}_N$ Markov inequality leads to
\[P_X\left (NH_N^2(P_X,Q)\geq \delta \text{ and }\forall R\in \mathcal{M}_N \;\; \log \frac{Q(X)}{R(X)}\geq 4(pen_N(Q)-pen_N(R))\right )\]
\begin{equation}\label{equ:markov}
\hspace{1cm} \leq \1_{NH_N^2(P_X,Q)\geq \delta}\E\left[\left (\frac{Q(X)}{R(X)}\right )^{1/4}\right ]e^{-pen_N(Q)+pen_N(R)}
\end{equation}

For all $R,Q\in \mathcal{M}_N$, by applying twice Cauchy-Schwartz inequality, we have:
\[\E\left[\left (\frac{Q(X)}{R(X)}\right )^{1/4}\right ] \leq \underbrace{\E\left[\left (\frac{Q(X)}{P_X(X)}\right )^{1/2}\right ]^{1/2}}_{A}\underbrace{\E\left[\left (\frac{P_X(X)}{R^+(X)}\right )\right ]^{1/4}}_{B}\underbrace{\E\left[\left (\frac{R^+(X)}{R(X)}\right )\right ]^{1/4}}_{C},\]
(this equation defines $A$, $B$ and $C$) where 
\[R^+=\Pi_{i=1}^N\left ( \pi_i(R)P_{i0}+(1-\pi_i(R))P_{i1}\right ).\]
 We first give an upper bound for $A$:
\begin{equation}\label{Abound}
A\leq e^{-\frac{N}{4}H_{N}^2(P_X,Q)}.
\end{equation}
This bound is easy to obtain by using the standard inequality 
\[\forall i=1,\dots,N\;\; \E_{P_{Xi}}\left[\left (\frac{Q_i(X[i])}{P_{Xi}(X[i])}\right )^{1/2}\right ]\leq e^{-\frac{h^2(Q_i,P_{Xi})}{2}}.\]
Equations (\ref{equ:markov}), (\ref{Abound}) and  (\ref{debu6}) and Assumption \ref{modelselc} (Kraft inequality) then give
\[\forall R\in \mathcal{M}_N,\; P_{X}(NH_N^2(P_X,\hat{P}_X)\geq \delta) \leq e^{pen(R)+\log\left (B(R)C(R)\right )}e^{-\frac{\delta}{4}}.\]
 We now only need to show that 
\begin{equation}\label{Bbound}
\log\left (B(R)\right )\leq c'h(R,N)
\end{equation}
($h(R,N)$ is given by Equation (\ref{gRN})) and 
\begin{equation}\label{Cbound}
\log\left (C(R)\right )\leq c'' \mathcal{L}_{N,n}
\end{equation}
with $c'>0$ and $c''>0$.\\

Let us begin with Equation (\ref{Bbound}). Easy calculous (using in particular the concavity of $x\rightarrow x^{1/2}$ lead to 
\begin{equation}\label{logtruc}
\log\left (B(R)\right )\leq \frac{1}{8}\sum_{i=1}^N\log\left (1+(\pi_i(P_X)-\pi_i(R))\E\left [\frac{P_{i0}-P_{i1}}{R_i}(X[i])\right ]\right )
\end{equation}
On the other hand,  Assumption \ref{BB} easily gives 
\[\left |\E\left [\frac{P_{i0}-P_{i1}}{R_i}(X[i])\right ]\right |\leq  \frac{2}{c}\]
(for a positive constant $c$) which gives, using Equation (\ref{logtruc}) and the inequality $\log(x+1)\leq x$ $\forall x>-1$, Equation (\ref{Bbound}).
\begin{Lemme}
Let $P$ and $Q$ be two equivalent probability measures. We then have
\[\sup_{x\in [0,1]}\E_Q\left [\frac{P}{xQ+(1-x)P}\right ]\leq 1\]
\[\sup_{x\in [0,1]}\E_Q\left [\frac{Q}{xQ+(1-x)P}\right ]\leq \max\left ( 1,\chi^2(Q,P)\right )+1\]
Let now $\tilde{P}$ and $\tilde{Q}$ be two other equivalent measures we then have
\[\sup_{x\in [0,1]}\E_Q\left [\frac{xQ+(1-x)P}{x\tilde{Q}+(1-x)\tilde{P}}\right ]\leq \max\left ( \chi^2(Q,\tilde{Q})+1,\E_Q\left [\frac{P-\tilde{P}}{\tilde{P}}\right ]+1\right ) \]
\end{Lemme}

The proof of this Lemma is only simple variational analysis (all the functions of $x$ that appear have maximum on $0$ or $1$), and the use of the identity
\[\E_{P}\left [\frac{P}{Q}\right ]=\chi^2(P,Q)+1.\]

We now show Equation (\ref{Cbound}). We have
\[
\log\left (C(R)\right )\leq \frac{1}{4}\sum_{i=1}^N\log\left (\E\left [\frac{\pi_i(R)P_{i0}+(1-\pi_i(R))P_{i1}}{\pi_i(R)\tilde{P}_{i0}+(1-\pi_i(R))\tilde{P}_{i1}}\right ]\right )
\]
and this, together with the third equation of the preceding Lemma, gives:
\[\log\left (C(R)\right )\leq \frac{1}{4}\sum_{i=1}^N\log\left (\left (\pi_i(P_X)D_{i0} +(1-\pi_i(P_X)) D_{i1}\right )+1\right ),\]
with, for all $i=1\dots,N$, 
\[D_{i0}=\max\left ( \chi^2(P_{i0},\tilde{P}_{i0}),\E_{P_{i0}}\left [\frac{P_{i1}-\tilde{P}_{i1}}{\tilde{P}_{i1}}\right ]\right )\]
and 
\[D_{i1}=\max\left ( \chi^2(P_{i1},\tilde{P}_{i1}), \E_{P_{i1}}\left [\frac{P_{i0}-\tilde{P}_{i0}}{\tilde{P}_{i0}}\right ] \right ).\]
Using the inequality $\max(a,b)\leq a+b$ for all $a,b\in \R$ gives the desired restult (i.e Equation (\ref{Cbound})).
 \end{proof}

\subsection{Proof of Theorem \ref{Mainth}}
The aim of this proof is to use Theorem \ref{expth}.
First, one should notice that Assumptions \ref{deux} and \ref{BB} imply assumption \ref{assump:low}. 
Let us define $\tilde{R}_{\pi}=(\tilde{R}_{\pi}[i])_{i=1,\dots,N}$ with 
\[\tilde{R}_{\pi}[i]=\frac{1-\hat{\pi}_i}{\hat{\pi}_i}\frac{dP_{i1}}{dP_{i0}}(X[i]).\] 
Then use the following lemma gives a kind of triangular inequality, it results from simple analysis.
\begin{Lemme}
There exits $c>0$ such that $\forall x,y\geq 0 $ 
\[\Omega(xy)\leq c\left (\Omega(x) +\Omega(y)\right ).\] 
\end{Lemme}
Using the Lemma with $x=R_{\pi}/\tilde{R}_{\pi}$ and $y=\tilde{R}_{\pi}/\hat{R}_{\pi}$ gives 
\[\Omega^2(R_{\pi}[i]/\hat{R}_{\pi}[i])\leq c \left ( \Omega^2(R_{\pi}[i]/\tilde{R}_{\pi}[i]) +\Omega^2(\tilde{R}_{\pi}[i]/\hat{R}_{\pi}[i])\right )\]
Also, because $P(X+Z\geq \delta)\leq P(X\geq \delta/2)+ P(X\geq \delta/2)$ for any real valued random variable $X,Y$, we have 

\begin{equation}\label{our:inequality}
 \forall \delta\geq 0  \;\; P_X\left ( \mathcal{E}(R_{\pi},\hat{R}_{\pi}) \geq \delta \right )\leq P_X\left ( \mathcal{E}(R_{\pi},\tilde{R}_{\pi}) \geq \delta/2 \right )+P_X\left ( \mathcal{E}(\tilde{R}_{\pi},\hat{R}_{\pi}) \geq \delta/2 \right ).
\end{equation}

We first bound the first term of the right hand side of Inequality \ref{our:inequality}.
Using Assumption \ref{BB} gives
\begin{align*}
\Omega^2(R_{\pi}[i]/\tilde{R}_{\pi}[i])&\leq c'|\pi(P_X)[i]-\hat{\pi}[i]|^2\\
& \text{ (for a positive constant }c'\text{ using Assumption \ref{BB}}) \\
& \leq c' \frac{h^{2}(P_{i0},P_{i1})}{|P_{i0}-P_{i1}|_1^2} \\
&\text{ from LeCam Inequality) }\\
& \leq c'' h^{2}(P_{i0},P_{i1}) \\
&\text{ from Assumption \ref{deux}). }
\end{align*}

Also, 

\begin{equation}
\sum_{i=1}^N\Omega^2(R_{\pi}[i]/\tilde{R}_{\pi}[i])\leq c'' N H_N(P_X,\hat{P}_X)
\end{equation}
and we can use Theorem \ref{expth} to conclude that 

\begin{equation}\label{parti:eq}
P_X\left (\mathcal{E}(R_{\pi},\hat{R}_{\pi})\geq \delta \right )`
\end{equation}
\[\;\;\leq \exp\left \{ c' \inf_{R\in \mathcal{M}_N}\{h(R,N)\}+c'' \mathcal{L}_{N,n} -\delta/4\right \},\]
where $h$  is given by Equation \ref{gRN}.\\

Let us now bound the second term of the right hand side of Inequality \ref{our:inequality}. To simplify notations, we set 
\[e=\sum_{i=1}^N\E\left [\Omega^2(R_{\pi}[i]/\tilde{R}_{\pi}[i])\right ]\]
The finite difference inequality implies that
\[\forall t\geq 0 \;\;  P\left ( \sum_{i=1}^N\Omega^2(R_{\pi}[i]/\tilde{R}_{\pi}[i])\geq e+t\right )\leq e^{-\frac{2t^2}{N}}\]
Also, taking $t=\left ( \delta/2-e\right )_+$ ( $(x)_+$ stands for the positive part of $x$) 
 gives
\[\forall \delta> 2e \;\;\; P\left ( \sum_{i=1}^N\Omega^2(R_{\pi}[i]/\tilde{R}_{\pi}[i])\geq \delta/2\right ) \leq e^{-\frac{2(\delta/2-e)^2}{N}}.  \]
Finally, because $e\leq N\phi_{N,M}$, there exists $c_0,c_1,c_2>0$ such that 
\[
 \forall \delta\geq 0  \;\;  P_X\left (\mathcal{E}(R_{\pi},\hat{R}_{\pi}) \geq \delta \right )\leq c_2e^{c_0N\phi_{N}-c_1\delta},
\]
 
 This inequality and inequality \ref{parti:eq} imply that Assumption \ref{assump:eta} from Theorem \ref{genth} is fulfilled and ends the proof.
\subsection{Proof of Theorem \ref{Th:last}}\label{Proof:LastTh}
First notice that the subscript $i$ in Theorem \ref{Th:last} does not play any role, we will chose $i=1$ and omit the corresponding subscript in the rest of the proof (in particular $g^*(x)$ will stand for $g^*(x)[1]$). Because $\mathcal{S}(\hat{g})$ is upper bounded by $1$, we only need to upper bound separately the $3$ terms that substantially appear in $\E[\min(\mathcal{L}_{N,n}/N,1)]$:
\[E_1=\E \left [\min \left (1,\E\left [\Omega^2\left (\frac{dP_{1}}{d\tilde{P}_{1}}\right ) \right] \right) \right],\;E_2=\E\left [ \min \left (1,\chi^2(P_{1},\tilde{P}_{1} ) \right) \right]\]
\[E_3=\E \left [\min \left (1,\E_{P_{0}}\left [\frac{P_{1}-\tilde{P}_{1}}{\tilde{P}_{1}}\right ] \right) \right]\]
where the last expectations in $E_1$ $E_2$ and $E_3$ are with respect to the learning set. 

{\bf Upper bound for $E_3$}
Simple calculation gives 
\begin{equation}\label{equ:E31}
\E_{P_{0}}\left [\frac{P_{1}-\tilde{P}_{1}}{\tilde{P}_{1}}\right ]=\E_{P_{0}}\left [e^{\mathcal{L}(X)}-1\right ]
\end{equation}
where 
\[\mathcal{L}(x)=\frac{1}{2}\left \langle C^-(\mu_1-\hat{\mu}_1),x-\frac{\hat{\mu}_1+\mu_1}{2}\right \rangle.\]
Setting $\xi=C^{-1/2}(X-\mu_0)$ in Equation \ref{equ:E31} gives 
\[\E_{P_{0}}\left [e^{\mathcal{L}(X)}-1\right ]=\E\left [e^{\tilde{\mathcal{L}}(\xi)}-1\right ]\]
where $\xi$ is a gaussian random variable with mean zero and covariance $I_p$ and $\tilde{\mathcal{L}}(x)=\mathcal{L}(C^{1/2}x+\mu_0)$. 
We now use corollary 1.7.9 in \cite{Bogachev:1998fk} which gives
\[\E\left [e^{\tilde{\mathcal{L}}(\xi)-\E[\tilde{\mathcal{L}}(\xi)]}\right ]\leq \E[e^{\|\nabla \tilde{\mathcal{L}}(\xi)\|^2_{\R^p}}]=e^{\|C^{-1/2}(\mu_1-\hat{\mu}_1)\|^2_{\R^p}} \]
and implies
\begin{align*}
\min(\E_{P_{0}}\left [e^{\mathcal{L}(X)}-1\right ],1)&\lesssim \|C^{-1/2}(\mu_1-\hat{\mu}_1)\|^2_{\R^p}+\left \langle C^{-}(\mu_1-\hat{\mu}_1),\mu_0-\frac{\hat{\mu}_1+\mu_1}{2} \right \rangle_{\R^p}\\
& \lesssim \|C^{-1/2}(\mu_1-\hat{\mu}_1)\|^2_{\R^p}+\left \langle C^{-}(\mu_1-\hat{\mu}_1),\mu_0-\mu_1 \right \rangle_{\R^p}
\end{align*}

Because we assumed that $C$ is known, we have 
\[C^{-1/2}(\mu_1-\hat{\mu}_1)=C^{-1/2}\mu_1-S_H(\xi)\]
where $S_H:\R^p\rightarrow \R^p$ is the hard threshold operator with threshold $\sqrt{2\frac{\log(p)}{n}}$ and $\xi$ is a	gaussian $\R^p$ random vector with mean $C^{-1/2}\mu_1$ and variance $\frac{1}{n}I_p$. Also, from Donoho and Johnstone \cite{DJ94a}, we can show that 
\[
\E\left [\min(\E_{P_{0}}\left [e^{\mathcal{L}(X)}-1\right ],1)\right ]\lesssim \sqrt{\frac{\log(p)}{n}}.
\]
{\bf Upper bound for $E_1$}
The upper bound for $E_1$ easily follows from the fact that 
\begin{align*}
\E\left [\Omega^2(dP_1/d\tilde{P}_1)\right ] \leq & \E \left [ \log^2\left ( dP_1/d\tilde{P}_1\right )\right ]\\
\leq & \E \left [ \mathcal{L}^2(X)\right ]
\end{align*}
{\bf Upper bound for $E_2$} follows directly from the upper bound for $E_3$ since 
\[\chi^2(P_1,\tilde{P}_1)=\E_{P_1}\left [e^{2\mathcal{L}(X)}\right ]-1.\]
This ends the proof. 
\subsection{Proof of corollary \ref{cro:birge}}
\begin{proof}
We only have to use Proposition 3  of Birgé \cite{Birge:2003rn}:
\begin{Lemme}
Let $Y$ be a positive random variable with 
\[P(Y>y)\leq \alpha e^{- y^2}\;\; \text{ for } y\geq \bar{y}\;\text{ and } \alpha>0.\]
Then, for all $q\geq 1$, 
\[\E[Y^q]\leq \bar{y}^q\left (1+\alpha\zeta_q(\bar{y})\right ),\]
where $\zeta_q$ is a function defined on $\R^+$ decreasing and such that
\[\forall x\geq cq, \;\; \zeta_q(x)=\frac{q}{2}e^{-x}, \;\text{ where } c=1/2 \;\text{ if }q\leq 2\pi e\;\;\text{ and }0.612 \text{ otherwise }.\]
\end{Lemme}
We applied the preceding Theorem to check the hypothesis of the Lemma with  
\[\bar{y}^2= 2 \left \{c' \inf_{R\in \mathcal{M}_N}\{g(R,N)\}+c'' \mathcal{L}_{N,n}\right \},\] $\alpha=e^{\frac{\bar{y}}{2}}$, $Y^2=NH^2_N(P_X,\hat{P})$. As a consequence, when $\bar{y}>(cq)^2$ we have
\[\alpha\zeta_q(\bar{y}) \leq  \frac{q}{2}e^{-\bar{y}/2},\]
which leads to the desired result (for $N$ large enough, and because $H_N^2\leq 2$, for all $N$ by changing the constant).
\end{proof}



\begin{thebibliography}{10}

\bibitem{Antoniadis:2009rt}
A.~Antoniadis, J.~Bigot, and R.~von Sachs.
\newblock A multiscale approach for statistical characterization of functional
  images.
\newblock {\em Journal of Computational and Graphical Statistics},
  18(1):216--237, 2009.

\bibitem{Barron:1999og}
A.~Barron, L.~Birg{\'e}, and P.~Massart.
\newblock Risk bound for model selection via penalization.
\newblock {\em probability theory and related field}, 113:301--413, 1999.

\bibitem{Birge:2003rn}
L.~Birg{\'e}.
\newblock Model selection via testing : an alternative to (penalized) maximum
  likelihood estimators.
\newblock {\em Annales de l'I.H.P. Probabilit{\'e}s et statistiques},
  42(3):273--325, 2006.

\bibitem{Bogachev:1998fk}
V.~I. Bogachev.
\newblock {\em Gaussian Measures}.
\newblock AMS, 1998.

\bibitem{esaimsurvey}
O~Bousquet, S~boucheron, and G~Lugosi.
\newblock Theory of classification: a survey of recent advances.
\newblock {\em ESAIM: Probability and Statistics}, 2004.

\bibitem{Devroye:1996fk}
L.~Devroye, L.~Gyorfi, and G.~Lugosi.
\newblock {\em A probabilistic theory of pattern recognition}.
\newblock Springer-Verlag, 1996.

\bibitem{wedg}
D~Donoho.
\newblock Wedgelets: Nearly-minimax estimation of edges.
\newblock {\em Annals of statistics}, pages 859--897, 1999.

\bibitem{DJ94a}
D~Donoho and Johnstone.
\newblock Ideal spatial adaptation by wavelet shrinkage.
\newblock {\em Biometrica}, 81(3):425--455, 1994.

\bibitem{mulgran}
E.~Kolaczyk, J.~Junchang, and S~Gopal.
\newblock Multiscale, multigranular statistical image segmentation.
\newblock {\em JASA}, 100(472):1358, December 2005.

\bibitem{multlik}
E.~Kolaczyk and R~Nowak.
\newblock Multiscale likelihood analysis and complexity penalized estimation.
\newblock {\em Annals of Stat}, 32(2):500--527, 2004.

\bibitem{imagerecons}
Korostelev and Tsybacov.
\newblock {\em Minimax Theory of Image Reconstruction}, volume~82 of {\em
  Lecture Notes In Statistics}.
\newblock Springer-Verlag, 1993.

\bibitem{Li:1999fk}
Q.~J. Li.
\newblock {\em Estimation of mixture Models}.
\newblock PhD thesis, Yale university, 1999.

\bibitem{Schmidt:2007sf}
F.~Schmidt.
\newblock {\em Classification de la surface de Mars par imagerie hyperspectrale
  OMEGA. Suivi spatio-temporel et {\'e}tudes des d{\'e}p{\^o}ts saisonniers de
  CO2 et H2O}.
\newblock PhD thesis, UJF, 2007.

\end{thebibliography}

\end{document}